%% file: From_Forecasting_Leaderboards_to_Deployment_Decisions_A_Fail_Closed_Certification_Protocol.tex
\documentclass{article}

\usepackage[margin=1in]{geometry}
\usepackage[T1]{fontenc}
\usepackage{microtype}
\usepackage{graphicx}
\usepackage{natbib}
\usepackage{hyperref}
\usepackage{amsmath,amssymb}
\usepackage{booktabs}
\usepackage{array}
\usepackage{tabularx}
\usepackage{threeparttable}
\usepackage{xcolor}
\usepackage{enumitem}
\usepackage{caption}
\usepackage{float}
\usepackage{adjustbox}
\usepackage[section]{placeins}
\usepackage{algorithm}
\usepackage{algorithmic}
\usepackage{tikz}
\usetikzlibrary{arrows.meta,positioning}
\hypersetup{hidelinks}
\setlist[enumerate]{topsep=0.35\baselineskip,itemsep=0.18\baselineskip,parsep=0pt,partopsep=0pt}
\newcolumntype{L}[1]{>{\raggedright\arraybackslash}p{#1}}
\newcolumntype{Y}{>{\raggedright\arraybackslash}X}
\newcommand{\githubmark}{\raisebox{-0.18em}{\includegraphics[height=1.05em]{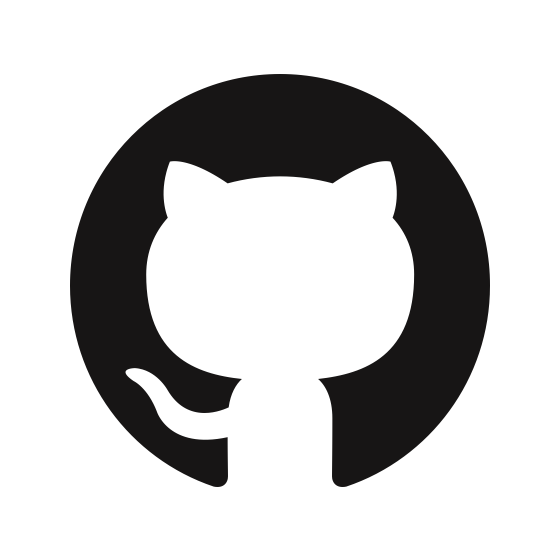}}}

\newcommand{\papertitle}{From Forecasting Leaderboards to Deployment Decisions: A Fail-Closed Certification Protocol}
\newcommand{\paperauthorblock}{%
\begin{tabular}{c}
\textbf{Geumyoung Kim}\\
Chungbuk National University\\
\texttt{goldzero@chungbuk.ac.kr}
\end{tabular}}
\newcommand{\makepapertitle}{%
\begingroup
\setlength{\parskip}{0pt}
\begin{center}
\vspace*{-0.45in}
\rule{0.78\textwidth}{2.4pt}\par
\vspace{1.25em}
\begin{minipage}{0.78\textwidth}
\centering\fontfamily{ptm}\fontsize{15.5}{18}\bfseries\selectfont
\papertitle\par
\end{minipage}\par
\vspace{1.05em}
\rule{0.78\textwidth}{0.55pt}\par
\vspace{2.05em}
\paperauthorblock
\end{center}
\vspace{1.25em}
\endgroup}

\begin{document}

\makepapertitle

\begin{center}
{\bfseries Abstract}
\end{center}
\vspace{-0.15em}
\noindent\hfill
\begin{minipage}{0.76\textwidth}
\small
\setlength{\parskip}{0.35\baselineskip}
Forecasting leaderboards rank models by predictive quality, but their winners are often read as deployment-ready top-1 advice. That reading can fail when forecasts are passed through a fixed decision interface, such as an alert threshold, a top-\(k\) budget, or a switching-cost policy. We study when a forecast-side winner can be certified as deployment-actionable for a specified interface and deployed utility. We introduce a fail-closed certification protocol whose gates are sufficient evidential conditions for a strong claim: a friction-caused, non-tie, statistically supported, and recurrent deployment-side reversal. Traffic-Hourly provides a certified anchor: winners agree at zero friction, but positive switching friction makes the forecast winner deployed-suboptimal. A locked native audit tests overclaiming: across 22 verified candidates and 362 full-grid cells, 155 apparent forecast/deployment winner inversions are blocked before certification. The contribution is not a new forecaster, metric, or universal utility, but a conservative protocol for deciding when forecasting leaderboard winners should be read as deployment-actionable top-1 advice.
\par\smallskip\begin{center}
\footnotesize\githubmark\ \textbf{GitHub:} \href{https://github.com/GamGomYang/forecast-actionability}{\texttt{github.com/GamGomYang/forecast-actionability}}
\end{center}
\end{minipage}
\hfill\null
\vskip 0.3in

\section{Introduction}

Forecasting leaderboards are usually designed to answer a predictive question: which model forecasts best under a chosen accuracy, calibration, or probabilistic scoring metric? In practice, however, a leaderboard winner is often used more strongly, as if it were also the safest model-selection recommendation for downstream deployment. This stronger interpretation can fail once forecasts are passed through a fixed decision interface. An alerting system may issue warnings when predicted risk exceeds a threshold, select the top-\(k\) alerts under a fixed budget, or penalize frequent action switches. A highly reactive forecaster can improve one-step predictive accuracy while inducing costly switching after the interface is applied. In such cases, the forecast-side winner need not be the deployed-side winner.

\begin{figure}[t]
\centering
\includegraphics[width=0.95\textwidth]{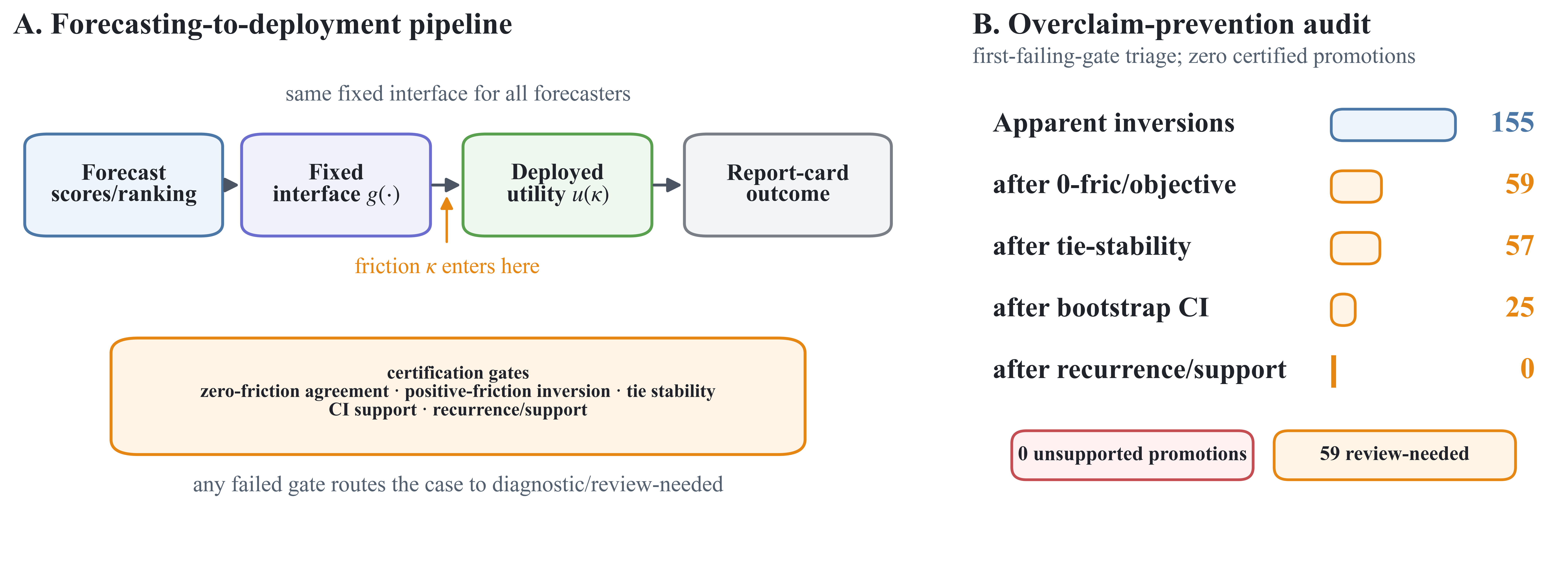}
\vspace{-0.62em}
\caption{Fail-closed certification for forecasting leaderboards. (A) A forecast-side winner becomes deployment-actionable top-1 advice only after a fixed interface, deployed-utility evaluation, and pre-specified certification gates. (B) In the locked native audit, 155 apparent forecast/deployment winner inversions are routed through the gates and zero receive certified promotion, illustrating overclaim prevention.}
\label{fig:fail_closed_report_card}
\vspace{-0.35em}
\end{figure}

This paper asks an audit question: when is a forecasting leaderboard winner sufficiently supported as deployment-actionable top-1 advice? The protocol does not replace the forecasting metric, retrain the forecasters, or propose a universal utility function. It audits a specified leaderboard cell together with a fixed forecast-to-decision interface and a deployed utility. The \emph{forecast-side winner} is the model ranked first by the forecasting metric, and the \emph{deployed-side winner} is the model with the highest deployed utility after the same interface is applied to every forecaster. Formally, for forecaster \(m\), let \(s_m\) denote its forecast-side score and \(u_m(\kappa)\) denote the deployed utility obtained after applying the fixed interface \(g\) under friction level \(\kappa\). The forecast-side winner is \(m_F=\arg\min_m s_m\) for a forecasting loss, whereas the deployed-side winner is \(m_D(\kappa)=\arg\max_m u_m(\kappa)\). A case is deployment-actionable only if the evidence supports reading \(m_F\) as reliable top-1 deployment advice under this specified interface.

The certification protocol is fail-closed: ambiguous cases are not promoted to headline failures. If the zero-friction baseline is not aligned, if the winner changes under a tie audit, or if the deployed shortfall lacks conservative uncertainty support, the row is routed to a diagnostic or review-needed outcome rather than certified as a deployment-facing selection failure. Each certification gate corresponds to a competing explanation for an apparent forecast/deployment winner mismatch: objective mismatch, absence of a positive-friction reversal, tie instability, statistical uncertainty, or insufficient recurrence/support. The report card is the user-facing output; the fail-closed certification protocol is the evidential rule that determines which rows, if any, may be promoted. The evaluated object is therefore not a standalone forecasting model, dataset, or scoring rule, but a leaderboard cell together with a specified forecast-to-decision interface and a pre-specified decision about whether its top-1 forecast winner is deployment-actionable. Appendix~\ref{app:report-card-definitions} gives the full label vocabulary and first-failing-gate rule.

The contribution is threefold: (i) a formalization of the deployment-facing interpretation problem for forecasting leaderboards, where the model that ranks first by forecast quality should not automatically be treated as the best deployment recommendation after a fixed interface; (ii) a fail-closed certification protocol that separates certified deployment-facing selection failures from objective mismatch, tie sensitivity, uncertainty limitation, low-support evidence, and no-detected-failure cases; and (iii) an empirical demonstration of both sides of the protocol, with Traffic-Hourly certifying a clean failure when all gates pass and the locked native audit blocking 155 apparent forecast/deployment winner inversions before promotion without sufficient evidence.

\FloatBarrier

\section{Results: Two Roles for the Certification Protocol}

\subsection{Evidence Roles: Anchor Certification vs. Overclaim Prevention}

We use the fail-closed certification protocol in two distinct evidence roles. The anchor suite asks whether the gates can certify a clean deployment-facing selection failure when the required assumptions hold. The locked native audit asks the complementary question: whether apparent forecast/deployment winner mismatches are prevented from becoming headline claims when the assumptions fail.

These roles are fixed before interpretation. Traffic-Hourly is the primary certified anchor; Event-micro is caveated support; the locked native audit is an overclaim-prevention audit; NOAA is frozen appendix confirmation; and inventory is a bounded operational check. This separation prevents mixed cases from being upgraded into primary positive evidence.

\begin{table}[!t]
\centering
\begin{threeparttable}
\caption{Clean anchor cases: Traffic-Hourly passes all certification gates under positive friction. The table reports whether the forecast-side winner becomes deployed-suboptimal after the fixed interface is evaluated.}
\label{tab:primary-selection-failure-anchors}
\vspace{-0.25em}
\scriptsize
\setlength{\tabcolsep}{2.0pt}
\input{results/table_primary_selection_failure_anchors.tex}
\begin{tablenotes}[flushleft]
\footnotesize
\item \emph{Note.} Mean shortfall is the paired deployed-utility loss from choosing the forecast-side winner instead of the deployed-side winner. Subopt. seeds counts the seeds where the forecast-side winner is deployed-suboptimal. Abbreviations: R-short = Reactive short, R-sharp = Reactive sharp, L-smooth = Lagged smoother, Calib. = Calibrated. Traffic family sweeps are in Appendix Table~\ref{tab:traffic-extended-interface-matrix}.
\end{tablenotes}
\end{threeparttable}
\vspace{-0.35em}
\end{table}

\subsection{Clean Anchor: Traffic-Hourly Can Be Certified}

Traffic-Hourly is a forecasting-native alert-selection setting: forecasters produce hourly risk scores, a fixed-budget interface selects the top-\(k\) alerts, and deployed utility rewards correct alert allocation while penalizing action switching through the friction parameter \(\kappa\). Table~\ref{tab:primary-selection-failure-anchors} isolates clean anchor cases from the broader native audit. In Traffic-Hourly, forecast and deployed winners agree at zero friction; under positive switching friction, Reactive short remains forecast-best but the deployed winner shifts to Lagged smoother or Calibrated/Lagged alternatives. In the representative Top-\(k\), \(k=249\), rows, the forecast-selected model is deployed-suboptimal in 100/100 seeds at both \(\kappa=0.5\) and \(\kappa=1.0\). Appendix Table~\ref{tab:traffic-extended-interface-matrix} reports the within-family breadth check over five Top-\(k\) budgets and two relative-rank variants.

The mechanism is intentionally transparent. The reactive model wins the forecast-side score because it tracks short-run variation, but this same reactivity induces costly switching after the fixed interface is applied. Once \(\kappa\) is positive, smoother alternatives can become deployment-optimal even though they do not win the forecast-side leaderboard.

Event-micro is retained as caveated support rather than a second primary anchor: its positive-friction rows are stable, but its near-zero row is less clean than Traffic-Hourly. It therefore supports the anchor pattern without being upgraded to primary evidence.

We use Traffic-Hourly as a mechanistic anchor rather than as a prevalence estimate. Thus, Traffic-Hourly is certified as a deployment-facing top-1 selection failure by the certification protocol: the zero-friction row rules out pure objective mismatch, and the positive-friction rows satisfy inversion, stability, uncertainty, and support requirements. Appendix checks cover Traffic-Hourly budgets, relative-rank variants, rolling-split proxies, and \(\epsilon\)-tie audits. Inventory remains a limited operational corroboration check rather than a third primary anchor.

\subsection{Locked Native Audit: Apparent Inversions Are Not Enough}

The pre-specified native audit tests the opposite risk: whether apparent forecast/deployment winner inversions would be promoted without sufficient evidence if reported naively. Here native means the locked real-data candidate grid used by the audit rather than the appendix-only NOAA check; locked means the candidate set and gate rules are fixed before assigning report-card labels. This audit is deliberately not used to estimate how often deployment-facing selection failures occur; it tests whether the certification protocol resists promoting naive winner mismatches when the certification assumptions are not met. The certification protocol is applied to 22 verified real-data native candidates and 362 native full-grid cells. Among these cells, 155 show an apparent forecast/deployment winner inversion. This audit is therefore adversarial to overclaiming rather than favorable to promotion: it starts from many apparent forecast/deployment winner mismatches and asks whether any survive the same evidential requirements imposed on the positive anchor.

Thus, the native audit provides an overclaim-prevention audit rather than a prevalence estimate: apparent forecast/deployment winner inversions are not promoted unless all fail-closed gates are satisfied. The zero-friction/objective gate blocks 96 rows, the tie-stability gate blocks 2, the bootstrap-CI gate blocks 32, and the recurrence/support gate blocks the remaining 25. No apparent forecast/deployment winner inversion in the locked native audit passes all gates. Thus, the audit yields zero certified promotions and zero failures promoted without sufficient evidence among 155 apparent forecast/deployment winner inversions. After the zero-friction/objective gate, 59 rows remain as review-needed rather than certified evidence.

This is the intended conservative behavior. The locked audit does not prove that deployment risk is absent; rather, it shows that winner mismatch alone is insufficient evidence for a deployment-facing selection-failure claim. Ambiguous rows remain visible as review-needed diagnostics instead of being converted into headline failures. Figure~\ref{fig:fail_closed_report_card}B visualizes the gate flow, and Table~\ref{tab:gate-ablation-overclaim} reports the exact ledger.

\input{results/table_gate_ablation_overclaim_audit.tex}

\subsection{Power and False-Promotion Diagnostics}

A certification-power diagnostic explains why low-support native rows are routed to review-needed: when already-certified Traffic-Hourly and Event-micro anchors are downsampled, no row is promoted at \(n \leq 20\), and promotion appears only around \(n=30\) or above.

This diagnostic explains why the support gate is not a cosmetic restriction. Under the same promotion logic, even already-certified anchor rows are not promoted when downsampled to very small support. Promotion appears only once the effective support is large enough for the direction and recurrence checks to become reliable. Thus, routing low-support native rows to review-needed is a deliberate fail-closed behavior rather than a failed detection.

Appendix~\ref{app:negative-control} reports the negative-control audit: 400 randomized winner assignments create many apparent forecast/deployment winner inversions, but no run clears the gates.

Finally, the frozen NOAA check is retained as appendix confirmation: it supports the high-friction pattern but remains mixed at lower friction, so it is not promoted to primary evidence (Appendix~\ref{app:noaa-confirmatory-rerun}).

These diagnostics support the interpretation that non-certification in the locked native audit is not merely a failure to find positive cases. The same rules also refuse to promote known positive anchors when support is too small, and they refuse to promote randomized negative controls despite many apparent forecast/deployment winner inversions.

\subsection{Why the Certification Gates Are Not Arbitrary}

The certification gates are not intended as necessary conditions for all possible deployment failures. They are sufficient conditions for promoting a strong paper-facing claim: a friction-caused, non-tie, statistically supported, and recurrent deployment-side reversal. This distinction is central to the fail-closed design. A row that fails a gate is not declared safe; it is routed to the first applicable diagnostic label.

The gates are ordered to remove competing explanations for an apparent forecast/deployment winner mismatch. The zero-friction gate removes pure objective mismatch: if the forecast-side winner and deployed-side winner already disagree at \(\kappa = 0\), then a positive-friction mismatch cannot be attributed to deployment friction alone. The positive-friction inversion gate then checks whether the forecast-side winner actually becomes deployed-suboptimal after the fixed interface and friction are applied. The tie-stability gate removes cases in which the reported winner identity is an artifact of an \(\epsilon\)-level ranking ambiguity. The conservative confidence-interval gate removes cases in which the paired deployed-utility shortfall is not statistically supported. Finally, the recurrence/support gate removes isolated seed, split, or grid artifacts that do not recur with sufficient pre-specified support.

This rationale is decision-theoretic rather than model-training based. Predict-then-optimize and decision-focused learning show that predictive quality and downstream decision quality need not be aligned, but those methods typically modify the training objective or optimize through a downstream problem. Our setting is post-hoc: forecasters, the leaderboard score, the forecast-to-decision interface, the deployed utility, and the friction grid are fixed before certification. The protocol therefore asks a narrower question: whether the top-1 forecast-side selection remains defensible as deployment-facing advice under that fixed evaluation object.

\paragraph{Proposition 1. Sufficient conditions for certifying a deployment-facing selection failure.}
Consider a fixed leaderboard cell with a fixed forecaster set, forecast-side score, forecast-to-decision interface, deployed utility, and friction level \(\kappa\). Let \(m_F\) denote the forecast-side winner and \(m_D(\kappa)\) denote the deployed-side winner after applying the fixed interface. A certified deployment-facing selection failure is promoted only if the following conditions hold:

\begin{enumerate}[label=(\roman*),leftmargin=2.4em,labelsep=0.55em,align=left]
\item \(m_F = m_D(0)\), so the forecast-side winner is also deployment-optimal before friction is introduced;
\item \(m_F \ne m_D(\kappa)\) for a positive friction level;
\item the reversal is not explained by an \(\epsilon\)-tie or unstable winner identity;
\item the paired deployed-utility shortfall from selecting \(m_F\) instead of \(m_D(\kappa)\) has a conservative confidence interval with positive lower bound; and
\item the reversal recurs with sufficient support across the pre-specified replicate or split structure.
\end{enumerate}

Under these conditions, the report card certifies a sufficient, not necessary, claim: the forecast-side leaderboard winner is not supported as deployment-actionable top-1 advice for the specified interface and utility. If any condition fails, the protocol does not infer deployment safety; it routes the row to the first applicable diagnostic label.

The locked native audit illustrates why the gates are needed. Starting from apparent forecast/deployment winner inversions alone would produce many apparent deployment warnings. After applying the same fail-closed certification sequence, however, no apparent native forecast/deployment winner inversion is certified as deployment-facing evidence. Thus, the audit does not show that deployment risk is absent; it shows that winner mismatch alone is insufficient for a deployment-facing selection-failure claim.

\FloatBarrier

\section{Discussion and Related Work}

\paragraph{Implications for forecasting leaderboards.}
Predictive rank and deployment-facing selection advice should be reported separately. A leaderboard can correctly identify the best forecaster under a forecast metric while still leaving open whether that winner remains best after a fixed decision interface. The report card adds an actionability layer to leaderboard cells likely to be used as deployment advice: it certifies only aligned zero-friction behavior, stable positive-friction inversion, uncertainty support, and sufficient recurrence, while objective-mismatch warnings and review-needed rows remain visible but not headline claims.

\paragraph{Decision-level forecast evaluation.}
Dynamic forecasting benchmarks such as ForecastBench and Prophet Arena evaluate forecasting capability on unresolved future questions and reduce contamination risk \cite{forecastbench,prophetarena}; time-series archives, foundation models, and benchmarks, including the Monash archive, GIFT-Eval, decoder-only TSFMs, Chronos, Moirai, text-conditioned forecasting, and TSFM evaluation pipelines, broaden the forecasters that may appear on future leaderboards \cite{godahewa2021monash,gifteval,das2024timesfm,ansari2024chronos,woo2024moirai,williams2025contextkey,tsfmbench2025,tempusbench2026,timerecipe2025}. Proper scoring rules remain central for probabilistic forecast evaluation \cite{gneiting2007,murphy1993}. Recent decision-level forecast evaluation work argues that forecasts should be evaluated not only by statistical accuracy but also by their value for downstream decisions. Weather and air-quality forecasting studies show that forecast-level model rankings can differ from decision-level rankings when the evaluation object includes a concrete decision task \cite{raeth2025,berlinghieri2024pm25}. Our setting is complementary: rather than designing a new decision-specific forecasting benchmark, we ask whether a given forecast-side leaderboard winner can be certified as deployment-actionable top-1 advice under a fixed interface and deployed utility.

\paragraph{Decision-focused learning and predict-then-optimize.}
Predict-then-optimize and decision-focused learning show that minimizing prediction error alone may be misaligned with downstream decision quality \cite{donti2017task,elmachtoub2021spo,mandi2024decisionfocused}. These methods typically modify training objectives or optimize through a downstream decision problem. Our protocol is post-hoc: it keeps the forecasters, forecast metric, fixed decision interface, deployed utility, and friction grid unchanged, and audits whether the top-1 forecast-side selection remains defensible as deployment-facing advice.

\paragraph{Benchmark uncertainty and rank instability.}
A separate line of benchmark work emphasizes that model rankings can be unstable under data sampling, initialization, hyperparameter choices, task aggregation, and statistical uncertainty \cite{bouthillier2021variance,longjohn2025aggregate,fevbench2025,brigato2025nochampions,neuhof2026rank}. This motivates the fail-closed use of tie audits, conservative confidence intervals, and recurrence/support requirements. In our setting, these checks are not cosmetic robustness tests; they determine whether an apparent forecast/deployment winner inversion is strong enough to be certified as a deployment-facing selection failure.

\paragraph{Positioning.}
Unlike decision-focused learning and predict-then-optimize, we do not train through the downstream decision. Unlike benchmark papers that primarily improve aggregate forecasting comparisons, we audit whether a particular forecast-side winner should be interpreted as deployment-facing top-1 advice after a fixed interface. The gate sequence is therefore a certification layer on top of an existing leaderboard cell, not a replacement for forecasting evaluation.

\paragraph{Limitations and scope.}
This paper should be read as an initial certification-protocol study, not as a comprehensive estimate of how often forecasting leaderboards fail as deployment advice. The clean certified evidence is concentrated in Traffic-Hourly. Event-micro provides caveated support, NOAA is retained as frozen appendix confirmation, and inventory is used only as a bounded operational check. Keeping these roles separate is important: mixed or lower-support cases are not upgraded into primary evidence.

The locked native audit has a similarly limited role. It is an overclaim-prevention audit, not evidence that deployment risk is absent. Its purpose is to show how apparent forecast/deployment winner inversions are blocked before certification when they are explained by objective mismatch, tie sensitivity, uncertainty limitation, or insufficient recurrence/support.

Finally, the protocol assumes fixed forecast-to-decision interfaces and fixed deployed utilities. It does not learn a new decision policy, propose a universal utility function, or replace forecast-side metrics. A natural next step is to evaluate the same fail-closed certification layer across larger pre-registered forecasting suites, adaptive decision interfaces, and online model-switching settings.

\clearpage
\begingroup
\small
\setlength{\emergencystretch}{2em}

\endgroup

\clearpage
\appendix

\setcounter{figure}{0}
\setcounter{table}{0}
\renewcommand{\thefigure}{A\arabic{figure}}
\renewcommand{\thetable}{A\arabic{table}}
\renewcommand{\theHfigure}{A\arabic{figure}}
\renewcommand{\theHtable}{A\arabic{table}}

\captionsetup{font=small,skip=4pt}

\providecommand{\appfigwidth}{0.92\linewidth}
\providecommand{\appwidefigwidth}{0.98\linewidth}
\providecommand{\apptablesize}{\small}
\providecommand{\apptighttablesize}{\footnotesize}

\setlength{\textfloatsep}{9pt plus 2pt minus 2pt}
\setlength{\floatsep}{8pt plus 2pt minus 2pt}
\setlength{\intextsep}{7pt plus 2pt minus 2pt}
\setlength{\tabcolsep}{4pt}
\renewcommand{\arraystretch}{1.06}

\section{Report-Card Output and Gate Procedure}
\label{app:report-card-definitions}

\paragraph{Appendix roadmap.}
Appendix A formalizes the report-card labels and first-failing-gate rule used by the fail-closed certification protocol. Appendix B fixes the evidence roles used in the paper. Appendix C reports positive-anchor robustness, Appendix D reports overclaim-prevention and false-promotion controls, and Appendix E collects closing checks and reproducibility notes. The appendix is therefore organized by evidential role rather than by experiment chronology.

\subsection{Label Vocabulary}

Table~\ref{tab:report-card-labels} gives the compact label vocabulary used by the report card. A certified deployment-facing selection failure requires zero-friction alignment, positive-friction winner inversion, stable tie behavior, conservative uncertainty support, and sufficient recurrence/support. Review-needed collects rows that show some warning signal but do not clear every certification gate. We use verified to mean that the cell passed the data-availability and fixed-interface replay checks required by the audit; it does not imply that the cell is certified as a selection failure.

\begin{table}[!htbp]
\centering
\caption{Report-card labels. The action column states how each outcome is used in the paper-facing evidence summary.}
\label{tab:report-card-labels}
\footnotesize
\setlength{\tabcolsep}{4pt}
\renewcommand{\arraystretch}{0.94}
\begin{adjustbox}{max width=\linewidth}
\begin{tabularx}{\linewidth}{@{}L{0.20\linewidth}L{0.49\linewidth}Y@{}}
\toprule
Label & Meaning & Action \\
\midrule
Deployment-facing selection failure & Forecast-side winner is deployed-suboptimal after all gates pass & Promote as certified evidence \\
Objective-mismatch warning & Forecast/deployed winners already differ at zero friction or zero row is missing & Block as non-friction-causal \\
Tie-sensitive & Winner identity or suboptimality changes under the tie audit & Route to review-needed \\
Uncertainty-limited & Directional deployed shortfall lacks conservative interval support & Route to review-needed \\
Excluded from evidence & Data or replay gates prevent use of the row & Exclude from evidence \\
No detected deployment-facing selection failure & No diagnostic found evidence against deployment-facing selection under the tested grid & Do not claim deployment safety \\
\bottomrule
\end{tabularx}
\end{adjustbox}
\end{table}

\subsection{Gate Rationale}

The certification gates are used as sufficient evidential conditions, not as necessary conditions for all possible deployment failures. Each gate removes a specific competing explanation for an apparent forecast/deployment winner mismatch.

\begin{table}[!htbp]
\centering
\caption{Gate rationale for the fail-closed certification protocol.}
\label{tab:gate-rationale}
\scriptsize
\setlength{\tabcolsep}{3pt}
\renewcommand{\arraystretch}{1.05}
\begin{adjustbox}{max width=\linewidth}
\begin{tabularx}{\linewidth}{@{}L{0.18\linewidth}L{0.25\linewidth}L{0.34\linewidth}Y@{}}
\toprule
Gate & Competing explanation removed & Required evidence & Failure routing \\
\midrule
Data/replay check & The row is not reproducible under the fixed interface & Forecasts, interface replay, deployed utility, and friction setting are materialized & Excluded from evidence \\
Zero-friction agreement & The mismatch is due to objective mismatch rather than friction & \(m_F = m_D(0)\) within the pre-specified tolerance & Objective-mismatch warning \\
Positive-friction inversion & No deployed-side reversal occurs after friction is introduced & \(m_F \ne m_D(\kappa)\) and the forecast-side winner is deployed-suboptimal at \(\kappa > 0\) & No detected deployment-facing selection failure \\
Tie stability & Winner identity is unstable under an \(\epsilon\)-level ambiguity & Winner identity and suboptimality survive the pre-specified tie audit & Tie-sensitive / review-needed \\
Conservative uncertainty support & The deployed shortfall is not statistically supported & Paired deployed-utility shortfall has a conservative confidence interval with positive lower bound & Uncertainty-limited / review-needed \\
Recurrence/support & The reversal is an isolated seed, split, or grid artifact & Direction and support thresholds are met across the pre-specified replicate or split structure & Review-needed \\
Certification & No competing explanation remains under the specified gates & All gates pass & Certified deployment-facing selection failure \\
\bottomrule
\end{tabularx}
\end{adjustbox}
\end{table}

This table also clarifies the interpretation of non-certification. A non-certified row is not evidence of deployment safety. It is evidence that the stronger claim of a friction-caused, non-tie, statistically supported, recurrent deployment-side reversal has not been established under the specified protocol.

\subsection{First-Failing-Gate Audit Procedure}

\floatname{algorithm}{Procedure}
Procedure~\ref{alg:fail-closed-promotion} states the report-card label assignment rule for one fixed-interface cell. This procedure is not intended as a learned algorithm. It is a reproducible audit rule: the first failed evidential condition determines the non-certified diagnostic label, and only rows passing all gates are promoted.

\begin{algorithm}[H]
\caption{First-failing-gate audit rule for certification-protocol labeling}
\label{alg:fail-closed-promotion}
\begin{algorithmic}[1]
\STATE Identify the forecast-side winner \(m_F\) under the forecast metric.
\STATE Identify the deployed-side winner \(m_D(\kappa)\) after applying the fixed interface.
\IF{data or replay checks fail}
    \RETURN \textsc{Excluded From Evidence}
\ENDIF
\IF{the zero-friction row is missing or \(m_F \neq m_D(0)\) beyond the pre-specified tolerance}
    \RETURN \textsc{Objective-Mismatch Warning}
\ENDIF
\IF{\(m_F = m_D(\kappa)\) at the tested positive-friction setting}
    \RETURN \textsc{No Detected Deployment-Facing Selection Failure}
\ENDIF
\IF{winner identity or suboptimality changes under the \(\epsilon\)-tie audit}
    \RETURN \textsc{Tie-Sensitive / Review-Needed}
\ENDIF
\IF{the paired deployed shortfall CI is not conservatively positive}
    \RETURN \textsc{Uncertainty-Limited / Review-Needed}
\ENDIF
\IF{recurrence or support thresholds are not met}
    \RETURN \textsc{Review-Needed}
\ENDIF
\RETURN \textsc{Certified Deployment-Facing Selection Failure}
\end{algorithmic}
\end{algorithm}

\FloatBarrier

\section{Data, Interfaces, and Candidate Families}
\label{app:data-interfaces-candidates}

The certification protocol is applied only after forecasts, fixed interfaces, deployed utilities, and replay checks are materialized. Table~\ref{tab:data-interface-summary} summarizes the evidence roles used in this paper; it is a scope map, not an expansion of the experimental claims.

\begin{table}[!htbp]
\centering
\caption{Data and interface scope. Evidence roles are kept separate: Traffic-Hourly is the primary anchor, Event-micro is caveated support, NOAA is frozen appendix confirmation, and inventory is a bounded operational check.}
\label{tab:data-interface-summary}
\footnotesize
\setlength{\tabcolsep}{3pt}
\begin{adjustbox}{max width=\linewidth}
\begin{tabularx}{\linewidth}{@{}L{0.18\linewidth}L{0.19\linewidth}L{0.23\linewidth}Y@{}}
\toprule
Setting & Evidence role & Fixed interface & Forecast-side criterion and units \\
\midrule
Traffic-Hourly & Primary anchor & Fixed-budget Top-\(k\) and relative-rank alert variants & Brier-style forecast-side loss over seed/replicate rows; deployed utility penalizes switching under \(\kappa\). \\
Event-micro & Caveated support & Fixed threshold and hysteresis checks & Brier is canonical, with log-loss reranking reported as robustness; 100 seed-level rows. \\
NOAA visibility & Frozen confirmation & Threshold-alert visibility warning & Frozen variants and split summaries are reported only in Appendix~\ref{app:noaa-confirmatory-rerun}. \\
Inventory & Bounded operational check & Fixed replenishment interface & Five-family operational proxy with mixed zero/low-friction behavior; not a primary anchor. \\
\bottomrule
\end{tabularx}
\end{adjustbox}
\end{table}

The native candidate audit uses naive-last, moving-average short/long, reactive-short, ridge-lag, and bridge aliases for the anchor families reactive-sharp, calibrated baseline, and lagged smoother. The fixed-interface families are threshold alert, top-\(k\) budget, hysteresis threshold, and capacity allocation. Alert-style grids use \(\kappa\in\{0,0.05,0.10,0.25,0.50,1.00\}\); allocation grids use \(\kappa\in\{0,0.001,0.005,0.010,0.050\}\). The certification-gate values and anchor robustness gates are defined in frozen configuration files.

\FloatBarrier

\section{Positive Anchor Robustness}
\label{app:positive-anchor-robustness}

\subsection{Traffic-Hourly Forecasting-Native Checks}
\label{app:traffic-redesign}

\subsubsection[Fixed-Budget Top-k Alert]{Fixed-Budget Top-\(k\) Alert}

\begin{table}[!htbp]
\centering
\apptighttablesize
\caption{Traffic-Hourly Top-\(k\) primary-anchor check under a fixed-budget interface. Agreement is strongest at zero friction and weakens at higher friction, where the deployed winner shifts away from the forecast-side winner.}
\label{tab:traffic-topk-alert-support}
\begin{adjustbox}{max width=\linewidth}
\input{results/table_q2_selection_drift_traffic_topk_main.tex}
\end{adjustbox}
\end{table}

In the selected Top-\(k\) setting, Reactive short is the forecast-side winner and the deployed winner at zero friction. At frictions \(0.5\) and \(1.0\), the deployed winner shifts to Lagged smoother even though Reactive short remains Brier-best and continues to win on the forecast side. The qualitative reading is the same as in the main text: under a fixed interface, the more reactive forecast-optimal family need not remain deployment-optimal once switching is penalized.

\subsubsection{Strict Extended Interface Matrix}

\begin{table}[!htbp]
\centering
\apptighttablesize
\caption{Traffic-Hourly extended interface matrix. All positive-friction rows retain the certification verdict across five Top-\(k\) budgets and two relative-rank variants under unchanged gates.}
\label{tab:traffic-extended-interface-matrix}
\begin{adjustbox}{max width=\linewidth}
\input{results/table_traffic_extended_interface_matrix.tex}
\end{adjustbox}
\end{table}

Table~\ref{tab:traffic-extended-interface-matrix} shows that the Traffic-Hourly certification pattern is stable across the tested Top-\(k\) budgets and relative-rank variants, supporting its role as the primary mechanistic anchor rather than a single selected cell.

\subsubsection{Relative-Rank Traffic Variant}

\begin{table}[!htbp]
\centering
\apptighttablesize
\caption{Traffic-Hourly relative-rank appendix variant for the primary anchor. The same qualitative separation between forecast-side and deployed-side selection appears under a relative-rank fixed-interface variant.}
\label{tab:traffic-relative-rank-support}
\begin{adjustbox}{max width=\linewidth}
\input{results/table_q2_selection_drift_traffic_relative_rank_support.tex}
\end{adjustbox}
\end{table}

As an additional Traffic variant, the relative-rank target shows the same strict certification pattern under a different forecasting-native label construction. Reactive short remains the forecast-side winner, while Calibrated baseline becomes the deployed winner once switching costs are introduced. The compact table shows \(m=0.10\); Table~\ref{tab:traffic-extended-interface-matrix} confirms that the alternate \(m=0.15\) variant also passes at both positive-friction levels. This breadth check stays within the existing clean Traffic-Hourly anchor rather than introducing a new domain claim.

\subsection{Event-Micro Robustness}
\label{app:event-micro}

Event-micro is retained as caveated supporting evidence. We construct a minimal forecasting-native binary-event setting evaluated over 100 seeds, in which each forecaster emits event probabilities, a fixed thresholding rule maps probabilities to actions, and switching friction penalizes action changes. The main text uses Brier as the canonical forecast-side criterion.

\begin{figure}[!htbp]
\centering
\begin{minipage}[c]{0.391\linewidth}
\centering
\includegraphics[width=\linewidth]{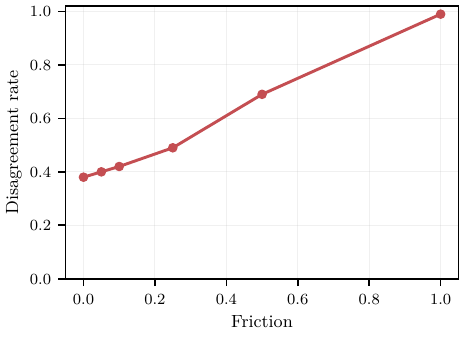}
\end{minipage}
\hspace{0.035\linewidth}
\begin{minipage}[c]{0.41\linewidth}
\centering
\footnotesize
\setlength{\tabcolsep}{3pt}
\begin{tabular}{ccccc}
\toprule
\(\kappa\) & Forecast & Deployed & Agree. & Shortfall \\
\midrule
0.00 & Sharp & Sharp & 0.62 & 0.003 \\
0.05 & Sharp & Sharp & 0.60 & 0.002 \\
0.10 & Sharp & Sharp & 0.58 & 0.002 \\
0.25 & Sharp & Calibrated & 0.51 & 0.003 \\
0.50 & Sharp & Calibrated & 0.31 & 0.011 \\
1.00 & Sharp & Smoother & 0.01 & 0.057 \\
\bottomrule
\end{tabular}
\end{minipage}
\caption{Event-micro caveated-support summary. Positive-friction rows show stable winner drift, but the weaker zero-friction behavior keeps Event-micro in a supporting rather than primary-anchor role.}
\label{fig:event-micro-canonical-summary}
\end{figure}

Log-loss reranking, alternate-threshold, and hysteresis-interface checks preserve the same qualitative winner-drift pattern, with smaller magnitudes under hysteresis. The compact interval table below records the interface-aware deployed-shortfall evidence without expanding the appendix into a full experiment log.

\begin{table}[!htbp]
\centering
\apptighttablesize
\caption{Interface-aware bootstrap intervals for the Event-micro caveated-support deployed shortfall. Positive intervals remain visible under both interfaces, with smaller shortfalls under hysteresis.}
\label{tab:event-micro-interface-gap-bootstrap-cis}
\begin{adjustbox}{max width=\linewidth}
\input{results/table_event_micro_interface_gap_bootstrap_cis.tex}
\end{adjustbox}
\end{table}

\subsection{Certification Power, Rolling Splits, and Tie Sensitivity}
\label{app:anchor-rolling}
\label{app:uncertainty-reporting}

These checks reuse the fixed anchor outputs and apply split-proxy gating over seed/replicate partitions, rather than expanding the broader real-data screen. Traffic-Hourly Top-\(k\) passes the zero-friction, positive-recurrence, bootstrap, and tie gates in all three split proxies. Event-micro remains supportive but carries the caveat stated in the main text: two split proxies pass the zero-friction gate, while one split is labeled objective-mismatch at zero friction and is therefore not promoted.

\input{results/table_anchor_downsampling_certification_power.tex}

\begin{table}[!htbp]
\centering
\apptighttablesize
\caption{Anchor-first rolling-split gate summary. Traffic-Hourly Top-\(k\) passes all split-proxy gates, while Event-micro remains caveated because one split does not clear the zero-friction gate.}
\label{tab:anchor-rolling-summary}
\begin{adjustbox}{max width=\linewidth}
\input{results/table_anchor_robustness.tex}
\end{adjustbox}
\end{table}

The uncertainty and \(\epsilon\)-tie checks are used only to prevent promotion without sufficient evidence. Rows that lack conservative interval support or have unstable winner identity are routed to review-needed rather than counted as certified selection failures. The tested relative tie thresholds, \(\epsilon_{\mathrm{rel}}\in\{0,0.001,0.005\}\), do not change the deployed-suboptimal conclusion or winner identity for the primary Traffic-Hourly and Event-micro positive-friction rows. Inventory tie checks remain part of the bounded inventory check and are not used to upgrade inventory to primary evidence.

Takeaway: the Traffic-Hourly anchor remains stable across the tested interface variants, while Event-micro remains supportive but caveated.

\FloatBarrier

\section{Native Overclaim-Prevention Controls}
\label{app:native-gate-ablation}

The locked native audit contains 22 verified candidates and 362 full-grid cells across bike share, electricity load, EPA air quality, retail demand, NYC taxi demand, weather alerts, and web traffic. This scope is used only for overclaim prevention: the audit asks whether apparent forecast/deployment winner mismatches survive the same certification gates required of the positive anchors.

Table~\ref{tab:gate-ablation-overclaim} in the main text can be read as an empirical gate-ablation ledger. Apparent forecast/deployment winner inversion alone would yield 155 warnings, but each gate removes a distinct competing explanation: objective mismatch, tie instability, uncertainty limitation, or insufficient recurrence/support. The final zero-promotion outcome is therefore not a null result; it is evidence that the protocol does not promote winner-identity mismatch into a certified deployment-facing claim without sufficient support.

\subsection{Negative-Control False-Promotion Audit}
\label{app:negative-control}

As a false-promotion check, we randomize winner identity in two ways: shuffling forecast winners and permuting deployed winners. These controls deliberately create many apparent forecast/deployment winner inversions while preserving the same gate logic used by the certification protocol. The purpose is to test whether apparent forecast/deployment winner mismatch alone can pass the certification gates; in 400 randomized runs, no run produces a gate-passing promotion.

\begin{table}[!htbp]
\caption{Negative-control audit. Randomized winners create many apparent forecast/deployment winner inversions; none clears all gates.}
\label{tab:negative-control-false-promotion-posthoc}
\centering
\scriptsize
\setlength{\tabcolsep}{2.0pt}
\begin{adjustbox}{max width=\linewidth}
\input{results/table_negative_control_false_promotion_posthoc.tex}
\end{adjustbox}
\end{table}

Takeaway: many apparent native forecast/deployment winner inversions are visible, but none survives the full certification sequence.

\FloatBarrier

\section{Appendix-Only Checks and Reproducibility}
\label{app:appendix-only-checks}

\subsection{Frozen NOAA Check}
\label{app:noaa-confirmatory-rerun}

The frozen NOAA check is retained as appendix confirmation. It supports the high-friction pattern but remains mixed at lower friction, so it is not promoted to primary evidence.

\begin{table}[!htbp]
\caption{Frozen NOAA visibility check. High friction confirms in two of three splits; lower friction remains mixed.}
\label{tab:noaa-confirmatory-rerun}
\centering
\scriptsize
\setlength{\tabcolsep}{2.5pt}
\begin{adjustbox}{max width=\linewidth}
\input{results/table_noaa_confirmatory_rerun_appendix.tex}
\end{adjustbox}
\end{table}

\subsection{Inventory Corroboration}
\label{app:inventory-corroboration}

Inventory is retained as a bounded operational check: the replenishment interface partly buffers forecast-side differences at zero and low friction, while higher friction favors smoother order trajectories, so the mixed pattern is not promoted to primary-anchor evidence.

\begin{table}[!htbp]
\centering
\apptighttablesize
\caption{Inventory check. Mixed zero/low-friction behavior keeps inventory in the bounded-corroboration role.}
\label{tab:inventory-selection-drift-appendix}
\begin{adjustbox}{max width=\linewidth}
\input{results/table_q2_selection_drift_inventory.tex}
\end{adjustbox}
\end{table}

\FloatBarrier

\end{document}

%% file: results/table_primary_selection_failure_anchors.tex
\begin{tabular}{@{}l c l l c r l l@{}}
\toprule
Setting & $\kappa$ & Forecast winner & Deployed winner & Subopt. seeds & Mean shortfall & 95\% CI & Evidence role \\
\midrule
Traffic Top-\(k\) \(k{=}249\) & 0.00 & R-short & R-short & 0/100 & 0.000 & [0.000,0.000] & zero-friction alignment \\
Traffic Top-\(k\) \(k{=}249\) & 0.50 & R-short & L-smooth & 100/100 & 7.016 & [6.756,7.283] & primary anchor \\
Traffic Top-\(k\) \(k{=}249\) & 1.00 & R-short & L-smooth & 100/100 & 24.651 & [24.226,25.079] & primary anchor \\
Event threshold \(\tau{=}0.55\) & 0.50 & R-sharp & Calib. & 69/100 & 0.011 & [0.009,0.014] & caveated support \\
Event threshold \(\tau{=}0.55\) & 1.00 & R-sharp & Smoother & 99/100 & 0.057 & [0.052,0.063] & caveated support \\
\bottomrule
\end{tabular}

%% file: results/table_gate_ablation_overclaim_audit.tex
\begin{table}[H]
\centering
\caption{Locked native overclaim-prevention ledger. No apparent inversion clears every gate.}
\label{tab:gate-ablation-overclaim}
\footnotesize
\setlength{\tabcolsep}{6pt}
\begin{tabularx}{0.63\linewidth}{@{}Yrr@{}}
\toprule
Gate / outcome & Remaining & Routed \\
\midrule
All native cells & 362 & -- \\
Apparent inversion screen & 155 & 207 \\
Zero-fric/objective gate & 59 & 96 \\
Tie-stability gate & 57 & 2 \\
Bootstrap-CI gate & 25 & 32 \\
Recurrence/support gate & 0 & 25 \\
\midrule
Final certified promotions & 0 & -- \\
Rows retained as review-needed & 59 & -- \\
\bottomrule
\end{tabularx}
\end{table}

%% file: results/table_q2_selection_drift_traffic_topk_main.tex
\begin{tabular}{lllllll}
\toprule
Friction & Forecast-side winner & Deployed winner & Agreement rate & Mean deployed gap & Median deployed gap & Deployed-suboptimal seeds / total \\
\midrule
0.00 & Reactive short & Reactive short & 1.00 & 0.000 & 0.000 & 0/100 \\
0.50 & Reactive short & Lagged smoother & 0.00 & 7.016 & 6.977 & 100/100 \\
1.00 & Reactive short & Lagged smoother & 0.00 & 24.651 & 24.473 & 100/100 \\
\bottomrule
\end{tabular}

%% file: results/table_traffic_extended_interface_matrix.tex
\begin{tabular}{llllllll}
\toprule
Interface & Param. & \(\kappa\) & Forecast winner & Deployed winner & Subopt. & Mean gap [95\% CI] & Verdict \\
\midrule
Top-\(k\) & \(k=125\) & 0.50 & Reactive short & Calibrated baseline & 100/100 & 3.794 [3.583, 4.010] & Pass \\
Top-\(k\) & \(k=125\) & 1.00 & Reactive short & Calibrated baseline & 100/100 & 15.986 [15.638, 16.341] & Pass \\
Top-\(k\) & \(k=186\) & 0.50 & Reactive short & Lagged smoother & 100/100 & 4.920 [4.675, 5.166] & Pass \\
Top-\(k\) & \(k=186\) & 1.00 & Reactive short & Lagged smoother & 100/100 & 20.281 [19.883, 20.678] & Pass \\
Top-\(k\) & \(k=249\) & 0.50 & Reactive short & Lagged smoother & 100/100 & 7.016 [6.756, 7.283] & Pass \\
Top-\(k\) & \(k=249\) & 1.00 & Reactive short & Lagged smoother & 100/100 & 24.651 [24.226, 25.079] & Pass \\
Top-\(k\) & \(k=312\) & 0.50 & Reactive short & Lagged smoother & 100/100 & 9.056 [8.795, 9.337] & Pass \\
Top-\(k\) & \(k=312\) & 1.00 & Reactive short & Lagged smoother & 100/100 & 28.832 [28.386, 29.295] & Pass \\
Top-\(k\) & \(k=500\) & 0.50 & Reactive short & Lagged smoother & 100/100 & 14.314 [14.056, 14.583] & Pass \\
Top-\(k\) & \(k=500\) & 1.00 & Reactive short & Lagged smoother & 100/100 & 37.985 [37.546, 38.447] & Pass \\
Relative-rank & \(m=0.10\) & 0.50 & Reactive short & Calibrated baseline & 100/100 & 10.103 [9.940, 10.271] & Pass \\
Relative-rank & \(m=0.10\) & 1.00 & Reactive short & Calibrated baseline & 100/100 & 25.205 [24.961, 25.445] & Pass \\
Relative-rank & \(m=0.15\) & 0.50 & Reactive short & Calibrated baseline & 100/100 & 16.328 [16.066, 16.612] & Pass \\
Relative-rank & \(m=0.15\) & 1.00 & Reactive short & Calibrated baseline & 100/100 & 35.581 [35.236, 35.930] & Pass \\
\bottomrule
\end{tabular}

%% file: results/table_q2_selection_drift_traffic_relative_rank_support.tex
\begin{tabular}{lllllll}
\toprule
Friction & Forecast-side winner & Deployed winner & Agreement rate & Mean deployed gap & Median deployed gap & Deployed-suboptimal seeds / total \\
\midrule
0.00 & Reactive short & Reactive short & 1.00 & 0.000 & 0.000 & 0/100 \\
0.50 & Reactive short & Calibrated baseline & 0.00 & 10.103 & 10.122 & 100/100 \\
1.00 & Reactive short & Calibrated baseline & 0.00 & 25.205 & 25.231 & 100/100 \\
\bottomrule
\end{tabular}

%% file: results/table_event_micro_interface_gap_bootstrap_cis.tex
\begin{tabular}{llllll}
\toprule
Interface & Friction & Mean deployed gap & Mean gap 95\% bootstrap CI & Median deployed gap & Median gap 95\% bootstrap CI \\
\midrule
Fixed threshold & 0.50 & 0.011 & [0.009, 0.014] & 0.008 & [0.004, 0.011] \\
Fixed threshold & 1.00 & 0.057 & [0.052, 0.063] & 0.053 & [0.048, 0.059] \\
Hysteresis threshold & 0.50 & 0.005 & [0.004, 0.006] & 0.001 & [0.000, 0.003] \\
Hysteresis threshold & 1.00 & 0.022 & [0.019, 0.026] & 0.020 & [0.015, 0.025] \\
\bottomrule
\end{tabular}

%% file: results/table_anchor_downsampling_certification_power.tex
\begin{table}[!htbp]
\centering
\caption{Downsampling diagnostic for certified anchor rows. Promotion appears only once support reaches roughly \(n=30\), illustrating why low-support native rows are routed to review-needed.}
\label{tab:anchor-downsampling-certification-power}
\scriptsize
\setlength{\tabcolsep}{4pt}
\begin{tabular}{@{}lccccccc@{}}
\toprule
Anchor & \(\kappa\) & \(n=6\) & \(n=12\) & \(n=20\) & \(n=30\) & \(n=50\) & \(n=100\) \\
\midrule
Event \(\tau=0.55\) & 0.50 & 0.000 & 0.000 & 0.000 & 0.927 & 0.985 & 1.000 \\
Event \(\tau=0.55\) & 1.00 & 0.000 & 0.000 & 0.000 & 1.000 & 1.000 & 1.000 \\
Traffic Top-\(k\) & 0.50 & 0.000 & 0.000 & 0.000 & 1.000 & 1.000 & 1.000 \\
Traffic Top-\(k\) & 1.00 & 0.000 & 0.000 & 0.000 & 1.000 & 1.000 & 1.000 \\
\bottomrule
\end{tabular}
\end{table}

%% file: results/table_anchor_robustness.tex
\begin{tabular}{llllll}
\toprule
Anchor & Zero gate & Positive recurrence & CI & Tie audit & Evidence role after split check \\
\midrule
Event-micro threshold & 2/3 & 3/3 & 3/3 & 3/3 & Caveated support \\
Traffic Top-\(k\) & 3/3 & 3/3 & 3/3 & 3/3 & Primary-anchor support \\
\bottomrule
\end{tabular}

%% file: results/table_negative_control_false_promotion_posthoc.tex
\begin{tabular}{lrrrrrrr}
\toprule
Control & Reps & Med. apparent & Max promo. & Med. promo. & Med. objective & Med. CI & Med. support \\
\midrule
Deployed-winner permutation & 200 & 275.5 & 0 & 0.0 & 74.0 & 183.0 & 19.0 \\
Forecast-winner shuffle & 200 & 276.0 & 0 & 0.0 & 75.0 & 183.0 & 19.0 \\
\bottomrule
\end{tabular}

%% file: results/table_noaa_confirmatory_rerun_appendix.tex
\begin{tabular}{lcccccl}
\toprule
Fixed variant & $\kappa$ & Zero align. & Cert. & Review & Other & Verdict \\
\midrule
q\_0.7 & 0.5 & 3/3 & 0/3 & 2/3 & 1/3 & review-needed/mixed \\
q\_0.7 & 1.0 & 3/3 & 2/3 & 0/3 & 1/3 & confirmed at $\kappa=1$ \\
q\_0.85 & 0.5 & 2/3 & 0/3 & 2/3 & 1/3 & review-needed/mixed \\
q\_0.85 & 1.0 & 2/3 & 2/3 & 0/3 & 1/3 & confirmed at $\kappa=1$ \\
\bottomrule
\end{tabular}

%% file: results/table_q2_selection_drift_inventory.tex
\begin{tabular}{llllll}
\toprule
Friction & Forecast-side winner & Deployed winner & Agreement rate & Mean deployed gap & Deployed-suboptimal seeds / total \\
\midrule
0.00 & Small MLP & Small MLP & 0.77 & 0.025 & 23/100 \\
0.25 & Small MLP & Small MLP & 0.76 & 0.039 & 24/100 \\
0.50 & Small MLP & Moving average (7) & 0.26 & 0.240 & 74/100 \\
1.00 & Small MLP & Moving average (7) & 0.01 & 1.029 & 99/100 \\
\bottomrule
\end{tabular}